\documentclass{article}

\usepackage[svgnames]{xcolor} 
\usepackage{graphicx}
\usepackage[margin= 0.8in,top=8mm]{geometry}
\usepackage{float}

\usepackage{verbatim}
\usepackage{algorithm}
\usepackage{algorithmic}

\newcommand*{\titleAT}{\begingroup 
\newlength{\drop} 
\drop=0.05\textheight 

\includegraphics[scale=1.5]{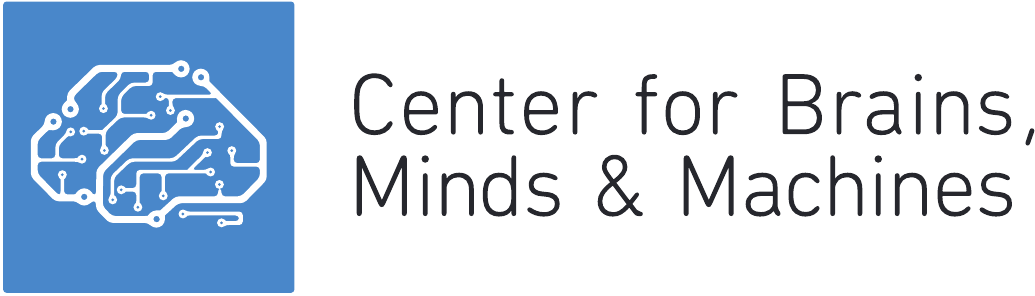}

\textcolor{CornflowerBlue}{\rule{\textwidth}{3 pt}}\par 
\vspace{2pt}\vspace{-\baselineskip} 
\rule{\textwidth}{0.4pt}\par 

\vspace{\drop} 
\textbf{\large{CBMM Memo No. \memonumber}}\quad \quad \quad\quad \quad \quad \quad\quad\quad \quad\quad\quad      \textbf{\large{\memodate}}

\vspace{\drop}
\begin{center}
\textbf{\huge{\memotitle}}\\
\vspace{0.4\drop}
\textbf{\Large{by}}\\
\vspace{0.4\drop}
\textbf{\large{\memoauthors}}
\end{center}
\vspace{\drop}
\textbf{\large{\noindent Abstract}:} {\memoabstract}

\textcolor{CornflowerBlue}{\rule{\textwidth}{3 pt}}\par 
\vspace{2pt}\vspace{-\baselineskip} 
\rule{\textwidth}{0.4pt}\par

\begin{minipage}{.15\linewidth}
\includegraphics[scale=0.1]{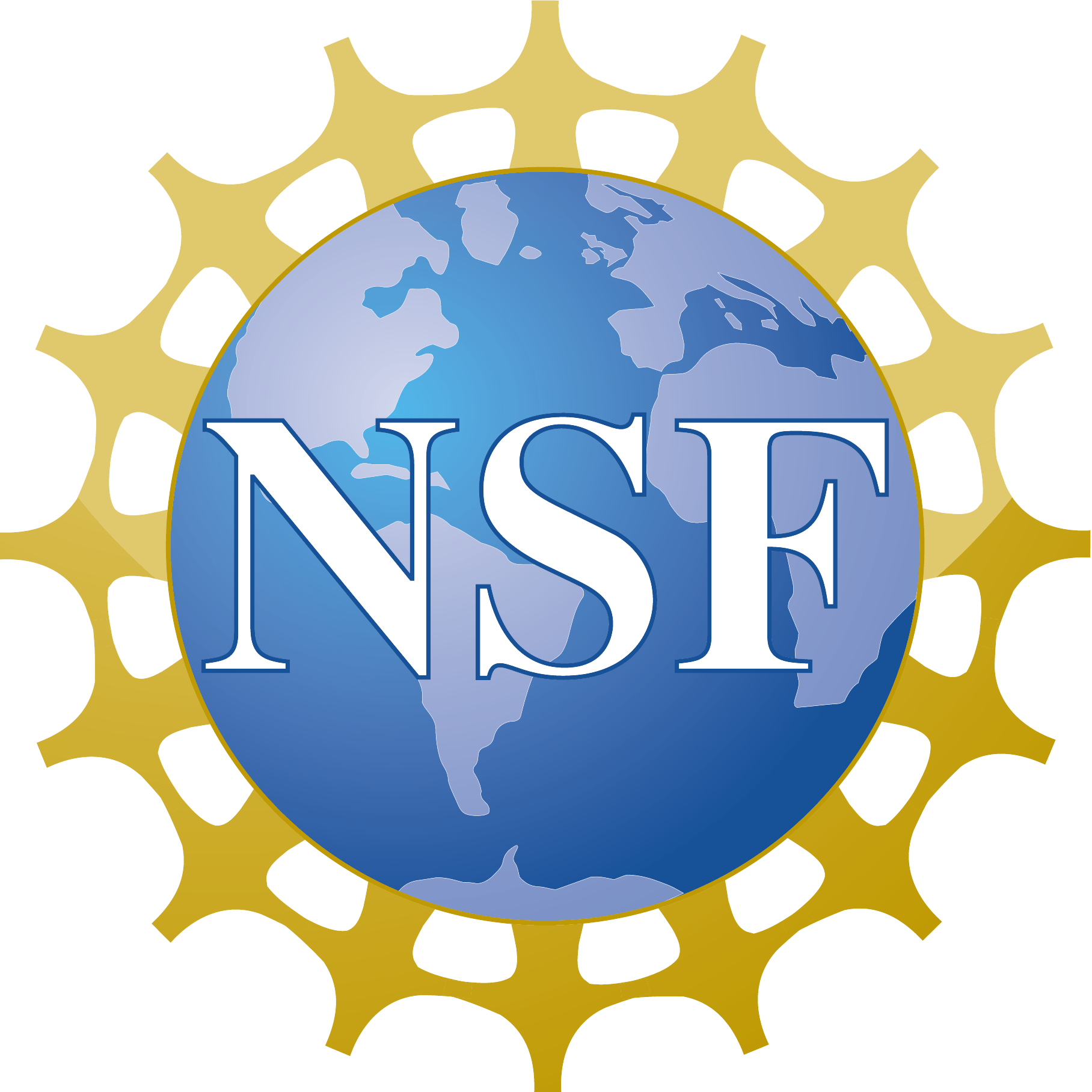}
\end{minipage}
\begin{minipage}{.84\linewidth}
\textbf{\large{This work was supported by the Center for Brains, Minds and Machines (CBMM), funded by NSF STC award  CCF - 1231216.}}
\end{minipage}
\endgroup}

\providecommand{\scal}[2]{\left\langle{#1},{#2}\right\rangle}
 \providecommand{\scalT}[2]{\left\langle{#1},{#2}\right\rangle}

\begin{document}

\pagestyle{empty} 

\def\memonumber{ 003 } 
\def\memodate{\today} 
\def\memotitle{Can a biologically-plausible hierarchy effectively replace face detection, alignment, and recognition pipelines?} 
\def\memoauthors{ Qianli Liao$^1$, Joel Z Leibo$^1$, Youssef Mroueh$^1$, Tomaso Poggio\footnote{MIT, McGovern Institute for Brain Research, Center for Brains, Minds and Machines} }
\def\memoabstract{The standard approach to unconstrained face recognition in natural photographs is via a detection, alignment, recognition pipeline. While that approach has achieved impressive results, there are several reasons to be dissatisfied with it, among them is its lack of biological plausibility. A recent theory of invariant recognition by feedforward hierarchical networks \cite{anselmi2013unsupervised}, like HMAX \cite{Riesenhuber1999,Serre2007a}, other convolutional networks (e.g., \cite{lecun1995convolutional}), or possibly the ventral stream, implies an alternative approach to unconstrained face recognition. This approach accomplishes detection and alignment implicitly by storing transformations of training images (called templates) rather than explicitly detecting and aligning faces at test time.   Here we propose a particular locality-sensitive hashing based voting  scheme which we call ``consensus of collisions'' and show that it can be used to approximate the full 3-layer hierarchy implied by the theory. The resulting end-to-end system for unconstrained face recognition operates on photographs of faces taken under natural conditions, e.g., Labeled Faces in the Wild (LFW) \cite{Huang2008}, without aligning or cropping them, as is normally done. It achieves a drastic improvement in the state of the art on this end-to-end task, reaching the same level of performance as the best systems operating on aligned, closely cropped images (no outside training data). It also performs well on two newer datasets, similar to LFW, but more difficult: LFW-jittered (new here) and SUFR-W \cite{Leibo}.}

\titleAT 

\newpage

\section{Introduction}

The challenge of simultaneously ensuring robustness to background clutter and invariance to appearance transformations is pervasive in the design of practical object recognition systems. One approach to mitigating both issues involves subjecting each test image to a detection, alignment, and recognition (DAR) pipeline. The DAR method handles clutter by detecting objects and cropping closely around them so that very little background remains. In a subsequent stage, it attempts to handle transformations by explicit alignment to a standard reference frame---typically effected by rotating and scaling the cropped images to bring certain key features into correspondence. DAR pipelines achieve impressive results in some cases. However, there may be no canonical way to align generic objects, especially if they are non-rigid, thus rendering the approach fundamentally limited.

DAR pipelines are also not plausible models of the brain's recognition system. The brain is not known to have any mechanism of alignment operating on the timescale of visual recognition. Moreover, in computer vision, the problem of unconstrained face recognition is perhaps the best case for the DAR approach since it is relatively simple to align faces to a canonical reference frame using correspondence of internal face features e.g., eyes, nose, and mouth. However, in Biology, face recognition is thought to be ``holistic''  \textit{i.e.,} not especially driven by the key features predicted by alignment-based strategies \cite{Tanaka1993}.

Here we investigate an alternative to the DAR approach based on the idea of using a feedforward hierarchy of stored templates to compute an invariant representation for new input images. This approach is both biologically plausible \cite{DiCarlo2012} and theoretically motivated \cite{anselmi2013unsupervised}. In order to compare with DAR pipelines, this paper focuses on the problem of unconstrained face recognition.  We show here that this approach yields an effective end-to-end system without explicit detection or alignment steps. In particular, we discuss how a system built according to the principles of a recent theory of invariance in hierarchical networks \cite{anselmi2013unsupervised} can evade the clutter problem---generally thought to be problematic for feedforward systems \cite{Treisman1988,itti2001computational,Chikkerur2010}. However, use of the system's basic version is limited by the time required to compute full convolutions over space and scale using a large number of filters (50,000 in this case). Thus, the other contribution of this paper is a locality sensitive hashing scheme combined with a voting scheme, that we call \emph{consensus of collisions} (CoC) that approximates the full system implied by the theory. This scheme allows faster computation and scalability of our system. 

We argue that the DAR framework guides researchers to focus on each stage of the pipeline in isolation, this is not negative \textit{per se}, but, through the development of specialized datasets researchers into the recognition subproblem come to focus on data that is biased toward the results of particular detection and alignment systems. In particular, Labeled Faces in the Wild-aligned (LFWa) \cite{Huang2008,Taigman2009}, the current gold standard dataset for the recognition step of unconstrained face recognition was filtered by the Viola-Jones face detector \cite{Viola2004} and consequently contains almost no faces with any significant rotation in depth. Similarly, alignment was accomplished by a particular commercial system \cite{Taigman2009} which likely introduces its own subtle biases. Leibo et al. (2014) \cite{Leibo} showed that these biases are severe enough that many recognition systems designed with LFWa in mind do not perform well on a new dataset (SUFR-W), gathered using a very similar protocol to LFW, the primary difference: substituting the, more tolerant of depth-rotation, Zhu \& Ramanan face detector \cite{Zhu2012} for Viola-Jones.

The system proposed here can take as input the full (unaligned, uncropped) images of the LFW dataset. Without performing explicit detection or alignment steps, it achieves a level of performance that compares favorably with the current state of the art systems operating on the aligned and cropped images ($87.55\%$). Note that the present system is solving a much harder problem. The accuracies quoted for the aligned and cropped LFW dataset assume perfect detection and alignment, whereas results using the full images include errors at those (implicit) steps. To further strain the system we also introduced a new ``jittered'' version of the LFW dataset with additional variation in face position, scale and orientation. The performance of classical systems such as HOG with SVM (Table \ref{table:jittered_performance}) drops significantly from LFW to LFW-jittered whereas the proposed system is largely unaffected by the additional transformations. We also show that this performance cannot be attributed to overfitting LFW, strong performance is still achieved when the system is trained on SUFR-W and tested on LFW (and vice-versa). 

These results demonstrate that the biologically plausible approach is not hopelessly stymied by clutter. Even in the case of unconstrained face recognition, where DAR pipelines could be considered most likely to be effective, this class of biologically plausible hierarchical networks are competitive with the current state of the art end-to-end systems. 

\section{Hierarchical architectures}
Hierarchical architectures alternating between selectivity-increasing ``tuning'' operations and tolerance-increasing ``pooling'' operations have a long history in computational neuroscience and computer vision \cite{Hubel1962,Fukushima1980,lecun1995convolutional,Riesenhuber1999}. Recently Anselmi et al. (2013) proposed a new theory of these architectures centered around the idea that they compute invariant representations called \emph{signatures}. Two previous face recognition models along the lines suggested by the theory have already been proposed---one was presented as a model of biology \cite{Leibo2011b}, the other as a face verification system \cite{Liao2013}. The present proposal incorporates ideas from both while significantly scaling up their scope of operation to the case of unconstrained face recognition without prior detection or alignment.


\begin{figure}[h]
\begin{center}
   \includegraphics[scale=.65]{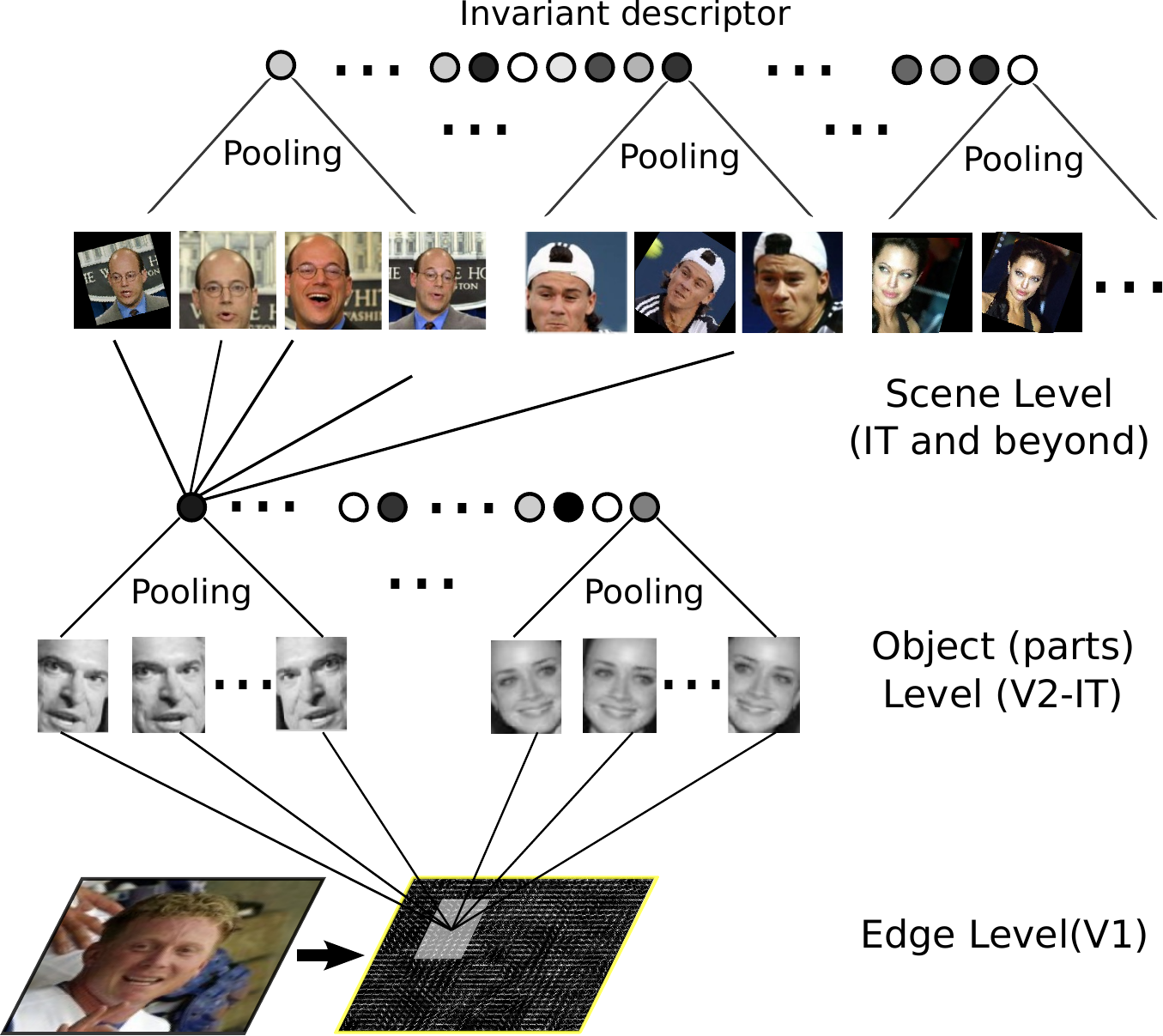}
\end{center}

   \caption{Illustration of the hierarchical architecture}
   \label{fig:architectur}
\end{figure}

\subsection{Theoretical motivation}

\noindent \textbf{Background: Invariance and discriminability}

Consider an image $I \in \mathcal{H}$ a Hilbert space. We are interested in recognizing the object depicted by $I$ even when it may have been transformed. Consider a family of transformations $G$ that may (or may not) be a group\footnote{For brevity, most of the exposition of the theory given here only applies to compact groups. See \cite{anselmi2013unsupervised} for the more general case.}. The orbit $O_I = \{gI |g\in G\}$ is itself an invariant with respect to the action of $G$. Notice that we use $g$ to refer both to an abstract element of $G$ and to its unitary representation acting on images. For example, if $G$ is the group of in-plane rotations, then the orbit is the set of images generated by  rotations of $I$. What if we had instead generated the orbit by transforming $I^\prime = \overline{g}I$? The two orbits are clearly identical, $O_I = O_{I^\prime}$.  That is, the set of all rotated images of $I$ is the same as the set of all rotated images of $\overline{g}I$. This motivates a definition. $I$ and $I^\prime$ are considered to be equivalent, written $I \sim I^\prime$,  when there exists a $g\in G$ such that $I = gI^\prime$. For example, $I$ and $I^\prime$ would be equivalent if they depict the same object from a different perspective. With this definition, it can be shown that if $G$ is a group then the orbit of any image $I$ under the action of $G$ is unique (see \cite{anselmi2013unsupervised}).

Next, let  $gI$ be a realization of a random variable. Consider the distribution $P_I$ of images obtained from $I$ under the action of $G$.  Anselmi et al. (2013) proved that, for $G$ a group, if two orbits coincide then their associated distributions under $G$ must be identical. This gives the following correspondence between images, orbits, and distributions.
\begin{equation}\label{eq:implications}
 I \sim I^\prime \iff O_I = O_{I^\prime} \iff P_I = P_{I^\prime}
\end{equation}
Thus the distribution $P_I$ is also invariant and unique to each object. The Cramer-Wold theorem \cite{Cramer1936} suggests a biologically plausible way to characterize such a distribution by its one-dimensional projections. A key result of the theory states (informally) that $P_I$ can be almost uniquely characterized by a set of $K$ one-dimensional distributions $P_{\scal{I}{t_k}}$ induced by the results of projecting $I$ onto a set of randomly chosen images $t_k, ~k = 1,\dots,K$ called \textit{templates}. The $P_{\scal{I}{t_k}}$ can themselves be characterized by their statistical moments, e.g., mean, max, etc. In practice, the number $K$ of projections needed to discriminate a finite number of orbits turns out not to be too large (and Anselmi et al. proved a bound \cite{anselmi2013unsupervised}).

Notice however, all this has shown so far is that \emph{if} you had stored (or could compute) $P_{\scal{I}{t_k}}$, \emph{then} the signature would be invariant and would discriminate between images of different objects. The key fact enabling this approach is that it is possible to store transformations of the templates instead.  That is, when $G$ is a group (and $g$ a unitary representation) then 
\begin{equation}\label{eq:transfer_condition}
\big\langle gI,t \big\rangle = \big\langle I,g^{-1}t \big\rangle
\end{equation}
This implies that the distributions $P_{\scal{gI}{t_k}}$ and $P_{\scal{I}{gt_k}}$ are identical. Thus, either one of them could be used to characterize $P_{I}$ and uniquely determine the associated orbit. The meaning of the result is as follows. It is not necessary to store all the transformations of $I$, or to have any explicit knowledge of $G$. Rather, storing transformations of the templates is sufficient to  construct an invariant for $I$ from a single view.

What happens in the non-group case? It can be shown that under fairly general conditions the signature will be \emph{approximately} invariant (see \cite{anselmi2013unsupervised}). However, in this case eq. \ref{eq:transfer_condition} no longer holds for all $t_k$. There is an additional requirement that $I$ and $t_k$ transform ``similarly'' (technically this is a condition on the tangent bundles of their respective orbits). Thus approximate invariance for non-group transformations is \emph{class-specific}. In order to compute an invariant signature of $I$ using stored templates in the non-group case, the object depicted in $I$ should be of the same class as the objects depicted in the $t_k$. For example, both $I$ and all the $t_k$ might be faces.  \cite{Leibo2011b} conjectured that this class-specificity is the reason the brain's ventral visual pathway separates the processing of faces from other objects \cite{kanwisher1997fusiform,Tsao2006}. \\

\begin{figure}[t]
\begin{center}
  \includegraphics[scale=.4]{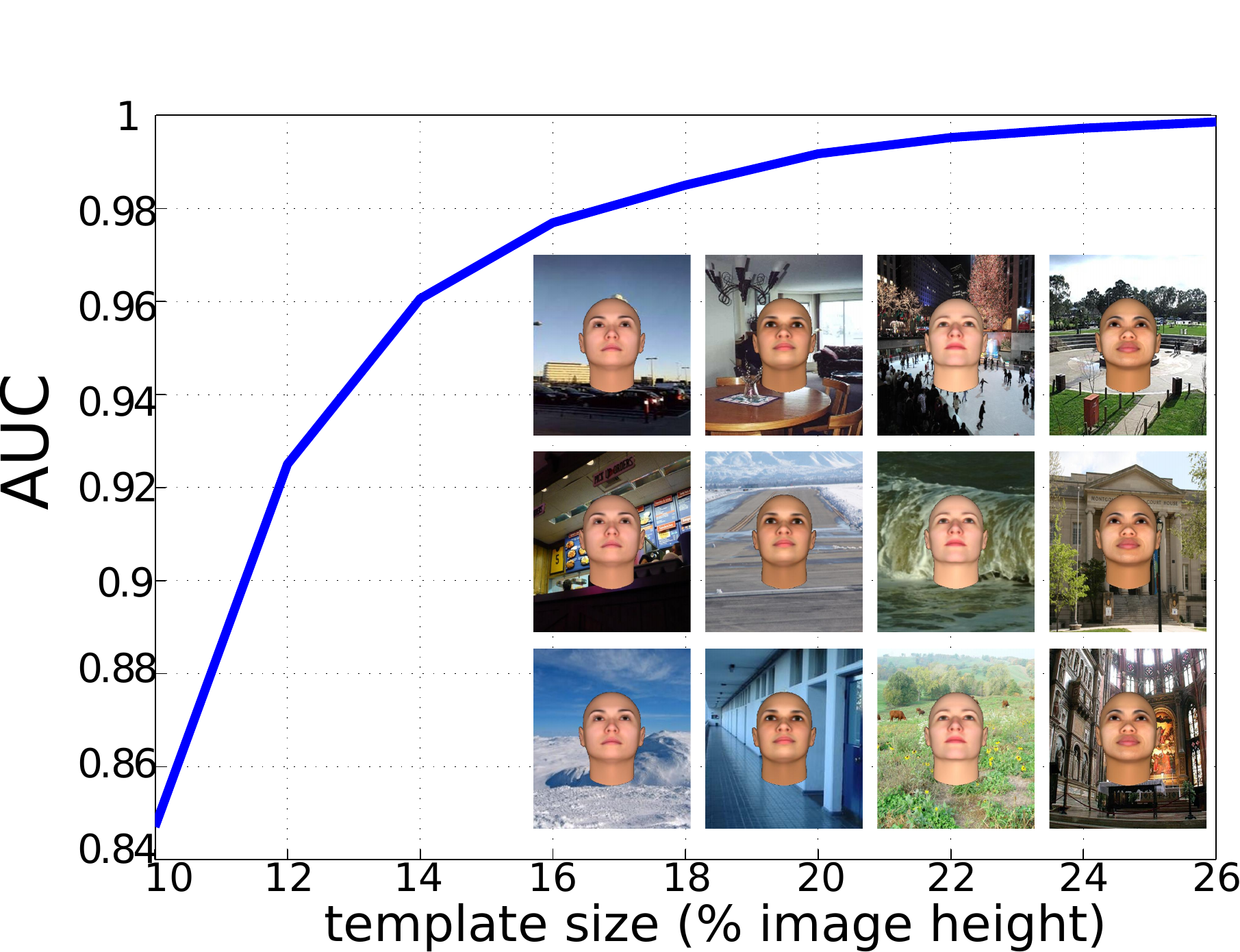}
\end{center}
   \caption{Evaluating clutter tolerance of globally translation invariant layer-2 signatures on a synthetic face verification (same-different matching) task. The only difference between different images of the same person is the background. The dataset is designed to have no other transformation, in order to isolate the clutter problem. The x-axis indicates the template size used, and y axis shows the AUC. Performance increases as templates get increasingly larger.  400 people and 5 background variation per person are used. 
}
\label{fig:synthetic_background}
\end{figure}

\noindent \textbf{Background: Architecture}

The architecture consists of a repeated biologically-plausible module that we call an HW-module in honor of Hubel and Wiesel's original proposal for the connectivity of V1 simple and complex cells \cite{Hubel1962}, which was extended as a hypothesis for other ventral stream areas in many computational models e.g., \cite{Fukushima1980,lecun1995convolutional,Riesenhuber1999}. Here we use the term HW-module to refer to a single ``C-unit'' and all its afferent ``S-units'' (note: these units need not correspond to actual cells). A layer of our architecture is a set of HW-modules. 

Each HW-module has a single template to which it is ``tuned''. The ``response''  $\mu^k(I)$ of the $k$-th HW-module to an image $I$ is given by
\begin{equation}\label{eq:HW-module}
  \mu^k(I) = P_g (\scal{I}{g t_k}).
\end{equation}
Where $P_g()$ is a pooling function. The two we will use in the present architecture are the mean and the max (over $g \in G$). 

Many different hierarchical architectures consisting of repeated HW-modules are possible.  Here we are primarily interested in architectures for which feature complexity as well as invariance increase from early to late layers. Some of these architectures seek to approximately ``factorize'' image variability into its component transformations. In such architectures, e.g., \cite{Leibo2011b,Liao2013}, smaller, edge-like templates are used in early layers while larger templates incorporating information from the entire image are used in higher layers. The early layers discount short-range group transformations while the higher layers compute a representation that is approximately invariant  to class-specific transformations. Anselmi et al. (2013) conjectured that these approximately factorizing architectures may improve the sample complexity of multistage learning since invariance in the lower layers could remove the need to align the training images for the higher levels. 

\subsubsection{The clutter problem\\}

A major challenge in the design of any face detection system is the problem of balancing the rate of target acceptance with the rate of false alarms on the  background. The hierarchical networks considered here do not have an explicit detection step but they do not escape these issues entirely. A translation-invariant HW-module tuned to a simple template will find high responses at many locations all over any natural image. Whereas a complex template will tend only to be activated when a part of the image is quite similar to it. To illustrate this, we constructed a ``pure'' test of clutter tolerance using synthetic face images. While these faces clearly do not capture the distribution of natural faces (e.g., none of them have hair), they are convenient for this demonstration. Since the faces themselves are fixed and only the background changes (\textit{i.e.} a model that was not translation invariant and only looked at the center of the image would always perform perfectly) the only way to fail is due to spurious activations on the background. Increasing the size of the templates steadily improves the performance on this pure test of clutter tolerance (fig. \ref{fig:synthetic_background}). Similar results with several other translation invariant architectures have been reported before, e.g., SIFT variants \cite{Ruiz-del-Solar2009,Leibo} so this is likely to be a general issue with translation invariance and not a quirk of our particular system or dataset.



\begin{algorithm}[h]
  \caption{Consensus of Collisions}
  \label{alg1}
  \begin{algorithmic}
    \STATE {\bfseries Input:} Test images  $I$, templates $\{t_j\}_{j=1\dots n}$, number of consensus $N$.\\
    \STATE {\bfseries Notations:}   $W:$ Oversampled windows, $C:$ Candidate Windows ,  $S: $ Consensus windows, $\scalT{}{}:$ normalized dot product.
    \STATE {\bfseries Output:} $R$ : the Response of templates for all images.
    \STATE {\bfseries Code:}
    \STATE {\bfseries Training:}
    \STATE Initialize hash function h 
    \STATE Compute hash codes of  $H_{t_j} =h(t_j)\quad  \forall j$ 
    \STATE  {\bfseries Testing:}
    \FOR {$i=1$ {\bfseries to} \texttt{\#images}}
    \STATE  W $\gets$ Dense oversampling of windows from image $I_i$
    \STATE Compute hash codes  $H_{w}=h(w),\quad  \forall w \in W $. 
    \FOR {$j=1$ {\bfseries to} \texttt{m}}
    \STATE $C_{t_j} \gets \{w \in W,| h(w)=h(t_j) \} $
    \ENDFOR
    \STATE $S\gets$  Select $N$ most frequently appearing windows out of $\cup_{j}C_{t_j}$
    \FOR {$j=1$ {\bfseries to} \texttt{n} }
    \STATE Set $P$ to be a empty Vector
    \FOR { all windows $w$ in $S$}
        \STATE $P(k) = \scalT{w}{t_j}$
        \STATE $k=k+1$
    \ENDFOR
    \STATE $R(i,j) = $\texttt{max}$(P)$
    \ENDFOR
    \ENDFOR
  \end{algorithmic}
\end{algorithm}

\section{Architecture and Approximations}\label{sec:arch}

Given a pair of images $(x_a,x_b)$ the task is to verify whether they depict the same person or not.  To test our HW-architecture we run it on both images and compare them by the angle between their signatures (top-level representations).  That is, we take the normalized dot product $\scalT{\mu(x_a)}{\mu(x_b)}$ if it exceeds  a threshold $\tau$, our method outputs that $x_a$ and $x_b$ have the same identity otherwise they are different. We use the training set to pick the optimal $\tau$.

In this section we detail our full architecture given in figure \ref{fig:architectur}. Since the architecture consists of a hierarchy of HW-modules, it can be thought of as a succession of simple and complex cells performing two main operations tuning (projection on a template) and pooling. The final output is the signature $\mu(I)$, a vector of top-level HW-module responses, each tuned to the identity of a face, invariantly to affine (group) transformations and approximately invariant to class-specific transformations.
\begin{enumerate}
\item \textbf{First Layer:} The proposed system uses closely cropped face images for training (but not for testing).  For each of the $n$ closely cropped face templates, we compute low-level features at each position/scale. These could be either HOG \cite{Dalal2005} features, or in some cases: HOG and LBP \cite{Ojala2002} features were combined. In those cases, the second layer was computed separately for each feature type and the results fused by concatenation before computing the third layer.. 
For training we extract $n$ HoG templates from \emph{closely cropped face images}. We call those templates the second layer training templates. 
\item \textbf{Second Layer:} We extract for each test image a dense overlapping set of $m$  windows.
We convolve the second layer training templates  with all the windows, and then apply max pooling over all scales and locations.
For each template, we pool the responses over all windows. Finally, for templates generated from a group of transformations (in-plane rotation in this case) our system also does max pooling over that group\footnote{Via eq. \ref{eq:transfer_condition}, for group transformations, transforming the test image or the template is equivalent.}.  
\item \textbf{Third Layer:} For each training image, run the architecture until the second layer. 
Store the responses up to the second layer and use them as the third-layer training templates. For a test image, compute the dot product of the output of the second layer, with the stored third layer training templates. Note that the third layer training templates are indexed by the identity of the person. Thus, as in \cite{Leibo2011b,Liao2013}, each third-level HW-module pools over a set of templates depicting the same person.   
\end{enumerate}


\begin{figure}[h!]
\begin{center}
  \includegraphics[scale=.65]{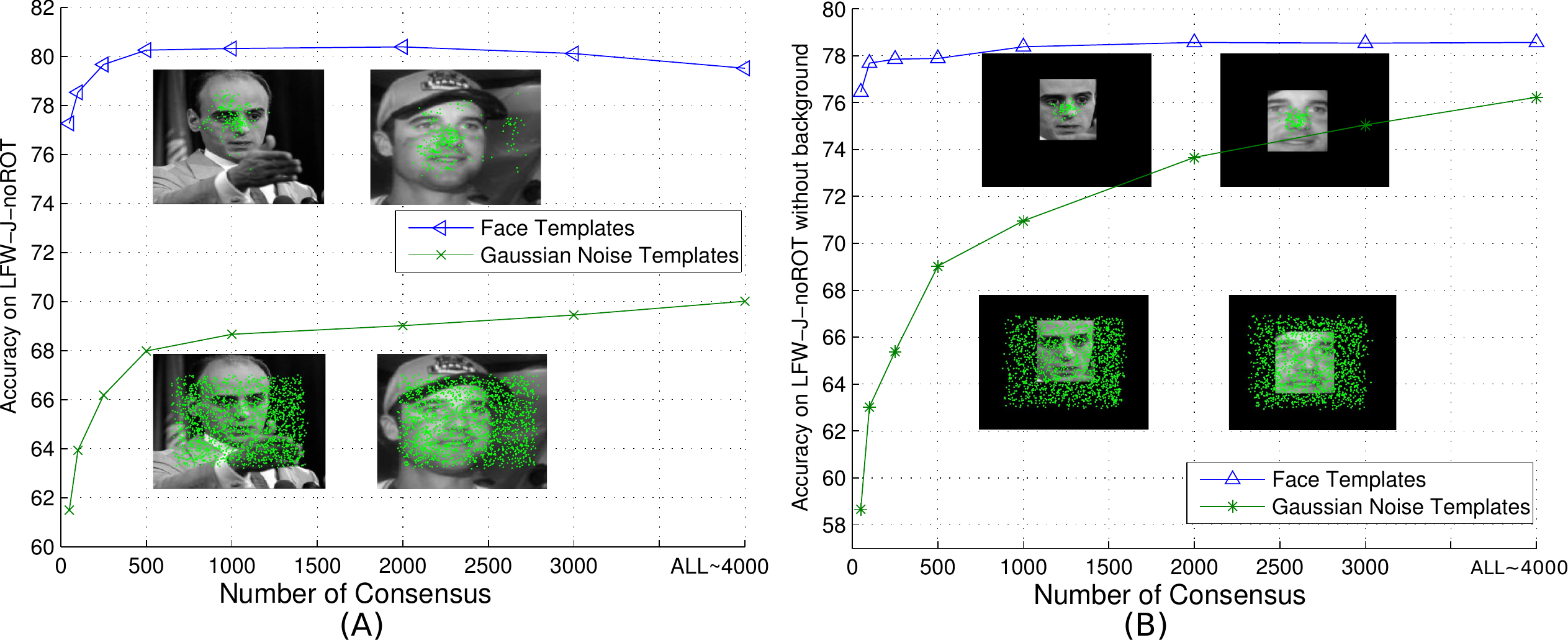}
\end{center}
   \caption{Evaluating the effect of ``Consensus of Collisions'' and background tolerance of class-specific templates. A set of experiments are run on two LFW-jittered-noRot datasets (for simplicity, no in-plane rotation jittering)---with or without background (figure A and B respectively). Each experiment used 5,000 templates of either faces or gaussian noise. The green dots in the images are examples of the maximum response locations of the templates of each curve (edges are blank because only valid convolution locations are considered). The last data point of each curve shows the performance when no CoC is performed. }
\label{fig:number_of_consensus}
\end{figure}

\subsection{Approximation by locality sensitive hashing and low rank approximation}
Layer two is the computational bottleneck. It is computationally / memory-access expensive to compute normalized dot products. Consider a set of windows $(w_1\dots w_m)$ representing an image by its sub-patches at a range of positions and scales. The problem is to compute the normalised dot product $\scal{w_i}{t_j}$ between each window $w_i$ and each template $t_j$ in a large collection.

Inspired by the impressive results of Dean et al. \cite{wta2013}, we experimented with approximating our architecture by locality sensitive hashing (LSH). Specifically, we propose a simple LSH scheme called ``Consensus of Collisions''  (CoC) (Algorithm \ref{alg1}) that evades the need for intensive memory accessing for dot product computations and reduces the computational load.  Let $h$ denote a randomised hash function. CoC proceeds in three steps in order to prune windows where it is unlikely to find a face:

\begin{enumerate}
\item \textbf{Hashing:} Produce hash tables of templates $(h(t_1)\dots h(t_n))$ and windows $(h(w_1)\dots h(w_m))$.
\item \textbf{Collect candidate ``match'' windows for each template:} For each template $t_j$, find a set of candidate windows that have exactly the same hash code. Let $C_{t_j}$ be that set.
\item \textbf{Voting Scheme:} For  each window $w_i$ compute the frequency of being chosen across templates, $f_{w_i}=\frac{1}{n}\sum_{j=1}^n1_{w_i \in C_{t_j}}$ . Pick the $N$ ``most popular" windows across templates i.e the $N$ windows having the highest frequencies. Let $S$ be the set of the $N$ most popular windows . We call $N$ \emph{the number of consensus}, and $S$ \emph {the consensus set}. 
 \end{enumerate}  

This locality sensitive hashing and voting scheme allow us to efficiently find a set of $N$ windows with high responses across templates. 
Note that $N \ll m$, so the speedup due to CoC is quite large. 

The final step is to pool across the windows of the consensus set $S$.  For each template $t_j$, find the window in $S$ with the highest response i.e., $\max_{w \in S }\frac{\scalT{t_j}{w}}{||t_j|| ||w||}$. We explored two options for computing the normalised dot products at this stage: 1. Exact computation, or 2., The PCA approximation described next.\\

\noindent \textbf{PCA approximation}

Note that the matrix $T$, consisting of all the templates (as column or row vectors), is a low rank matrix. We can perform PCA on $T$ and keep the $k$ largest eigenvectors. By projecting the templates and the windows to the $k$ dimensional space defined by those eigenvectors we can perform faster dot products in the reduced dimensional space. This procedure is similar to the one employed by \cite{chikkerur2011approximations}. 

%
%


\section{Experimental Evaluation}
In this section, we demonstrate the performance and properties of our
approach through a set of experiments on LFW, SUFR-W and a series of
difficult LFW-jittered datasets we created.

\subsection{Datasets}
\begin{figure}[t]
\begin{center}
  \includegraphics[scale=.35]{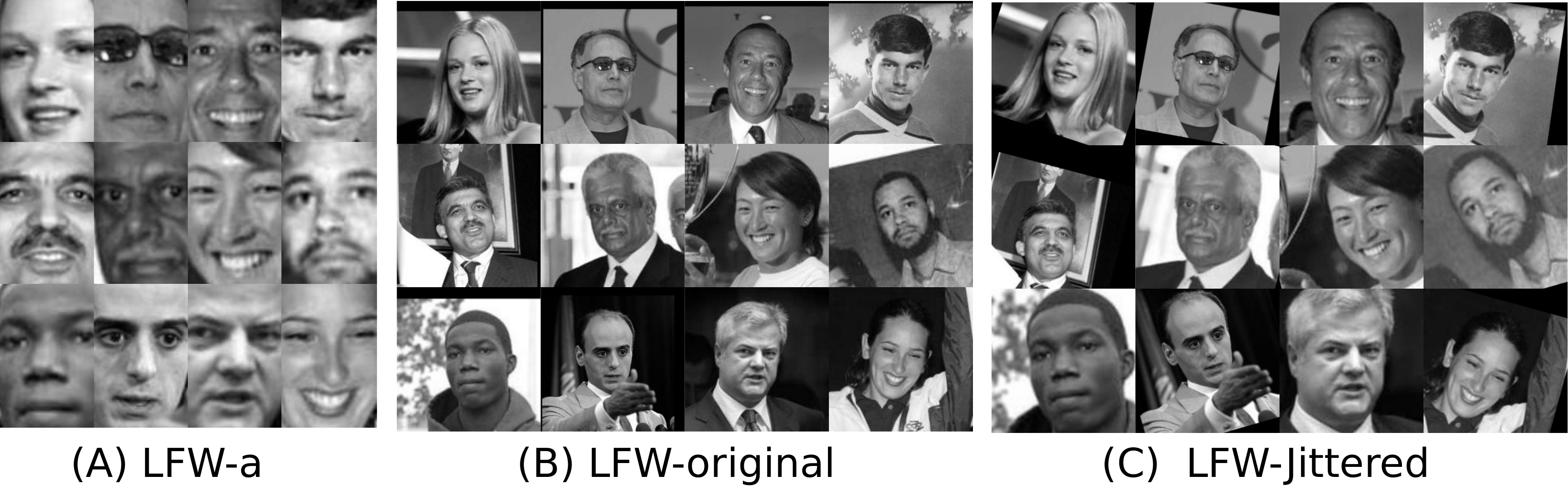}
\end{center}
   \caption{(A) LFW-a dataset---The majority of studies on LFW actually use the more finely aligned dataset LFW-a and crop very close as shown here  \cite{Leibo}. (B)  Original LFW images. (C) Our LFW-jittered dataset.}
\label{fig:gallery}
\end{figure}

We used the following five datasets throughout the paper:(1) LFW-original: The full (250x250), non-cropped, non-aligned version of LFW dataset (Figure. \ref{fig:gallery}). (2) SUFR-W: Unconstrained face recognition dataset collected by \cite{Leibo} with similar protocol to LFW but with a more advanced detector \cite{Zhu2012}. (3) LFW-Jittered (LFW-J): The LFW-Jittered dataset was created by randomly translating, scaling and in-plane rotating LFW original images. Translation range: -40 to 40 pixels, scaling range: 1 to 1.5, in-plane rotation range: -20 to 20 degrees. See the supplementary information for more details on the datasets and example images.

\subsection{Experiments}

\noindent \textbf{Full model:} We tested the full model described in Figure \ref{fig:architectur} on the LFW-original dataset without using any hashing approximation (first two rows of Table \ref{table:speedup}).\\ 

\noindent \textbf{Approximated model:}  Despite the encouraging accuracy of the direct implementation, it is not fast enough for practical purposes. Thus, we further explored the effect of hashing and several different choices of PCA and feature type (the rest of table \ref{table:speedup}). The system achieved 82.53\% performance on the original LFW data with a system that runs at nearly 2 frames per second.  
Note that hashing becomes indispensable when the windows number becomes large. We tested a model with a large number of scales---the  pyramid contains 31-scales from 125x125 (50\%) to 500x500 (200\%), generating about 30,000 windows at test time. In this case, any model without hashing is memory-intractable (requires $>$25 GB per thread), and hashing alone gives a 60x speedup.\\

\noindent \textbf{Large class-specific templates:} As suggested by fig. \ref{fig:synthetic_background}, large templates mitigate the clutter problem of translation-invariant
HW-modules. To experimentally demonstrate this property, a set of experiments were run on LFW-jittered-noRot and
LFW-jittered-noRot-noBG. For each experiment, we used  5,000 templates of either faces or gaussian noise. Performances are plotted as a function of the number of consensus $N$ kept (Figure \ref{fig:number_of_consensus}). The performance of gaussian noise templates is surprisingly high when no clutter is present, but much lower otherwise. With face templates, the model tolerates significant clutter and is highly selective to faces, in which case hashing is able to reduce the number of windows computed by 90\% without lowering the performance.\\

\noindent \textbf{Robustness across datasets:}
Despite the recent close-to-human performance reported in LFW-a, (e.g., \cite{Chen2013}), it has been argued by \cite{Leibo} that good LFW performance often does not transfer to SUFR-W, indicating that the community may be somewhat overfitting LFW. To demonstrate the robustness of our model, we tested its performance across LFW and SUFR-W. The approach is to train on one dataset and test on the other. Since our model has two stages of trainable templates, we tried either partial or full training using the other dataset's training set. The performance is shown in Table \ref{fig:dataset_bias}. The findings are: 1. training the first layer on either dataset gives similar performance.     2. LFW is significantly easier than SUFR-W. \\

\textbf{The LFW-original state-of-the-art:}
In Table \ref{table:final_performance}, we demonstrate our system's performance on LFW, comparing to the ``no outside data used, unaligned'' category. Note however, since our system uses the identities of the faces in the training set, it does not exactly conform to the recommendations of \cite{Huang2008} for ``restricted'' training with LFW which requires only same-different pairs to be used. Our procedure follows the ``restricted" protocol at test time, but not for training. We intend to model what the brain could learn via unsupervised mechanisms---these template orbits could have been learned in an unsupervised fashion by observing faces transform and pooling over temporally adjacent frames. The same setting is addressed in \cite{Leibo2011b,Liao2013,anselmi2013unsupervised}.\\

\noindent \textbf{Jittering Invariance:}
Human vision is invariant to small shifts in object position. Motivated by the strong performance on LFW-original, we tested the same model on LFW-jittered data (Table \ref{table:jittered_performance}), which we expected to be very difficult for conventional computer vision methods. Our system achieved almost the same performance on this dataset. In contrast, the baseline model (HOG) drops by almost 20\%. Intriguingly, our training templates were only rotated between -12 to 12 degrees, but the model handles -20 to 20 jittering in LFW-J without any problem. This is likely due to preexisting angle variation in the training templates (they were not aligned). As in \cite{Liao2013}, Non-uniform and ultra-sparse sampling of the orbit are sufficient for good performance \cite{Liao2013}.\\



\begin{table*}[t]
 \begin{center}
  \begin{tabular}{|ccc|cccc|}
 \hline  
{\bf Feature}  & {\bf \#EigenVec.}  &  {\bf \#Consensus } & {\bf Mem.} & {\bf Time.}   & {\bf Acc.}  & {\bf Speedup}  \\
\hline
LBP & No PCA  & No Hashing  & 6.8 GB & 44.67 & 83.67 & 1x \\
LBP & No PCA   & 500  & 2.9 GB & 4.38 &  82.40 & 10.2x \\
LBP & 250   & No Hashing  & 2.1 GB & 4.13 & 82.18 & 10.8x  \\
LBP & 1200  & 1500  & 2.0GB & 2.51 &  83.17  & 17.7x \\
HOG & 1200  & 1500  & 1.8GB & 1.69 &  {\bf 84.73}  & 26.4x \\
LBP & 250   & 500  & 1.5 GB & 0.89 &  81.18 & 50.2x \\
HOG & 250   & 500  & 1.1 GB & {\bf 0.54} &  82.53 & {\bf 82.7x} \\
 \hline
\end{tabular}
\caption{ The performance (Acc.) evaluated on the unaligned LFW dataset. There were about 50,000 face templates in our
  model. The testing image (250x250) was scaled to form a 12-scale
  pyramid with size ranging from 288x288 to 150x150. The dimensionalities of the
  LBP \cite{Ojala2002} and HOG \cite{Dalal2005} features were 7540 and 4030 respectively. Every experiment
  was run on a single 2010 machine with an 8-core processor and 36GB
  RAM. No GPU was used. Multi-threading was employed to use the CPU as
  much as possible. The time refers to the average time spent on a
  single frame (13233 frames are tested in total). The ``Mem'' refers
  to the average memory requirement of each thread. The code was
  written in Matlab and still has optimization potential. A GPU
  implementation was found to be inefficient due to the GPU's small on-board
  memory. The number of CPU cores is the bottleneck. Further speedup is expected if more of them are available. Hash code length: 24,
  Hash table number: 20. Note: The training process just performs PCA and stores templates---it only takes about 5 to 10 minutes (contrast with most deep learning approaches (e. g., \cite{lecun1995convolutional}). }
\label{table:speedup}
\end{center}
\end{table*}
\vskip -0.42in

\begin{table}[t]
 \begin{center}
  \begin{tabular}{|ccc|c|}
 \hline
{\bf 2nd Layer}  & {\bf 3rd Layer}  &  {\bf Test on } & {\bf Acc.} \\
 \hline
SUFR-W & SUFR-W & SUFR-W  & 80.30$\pm$0.89\%   \\
LFW & SUFR-W & SUFR-W  & 79.60$\pm$1.41\%   \\
LFW & LFW & SUFR-W  & 76.08$\pm$1.36\%   \\
 \hline
LFW & LFW & LFW  & 84.33$\pm$1.79\%   \\
SUFR-W & LFW & LFW  & 83.87$\pm$1.38\%   \\
SUFR-W & SUFR-W & LFW  & 84.55$\pm$1.43\%  \\
 \hline
\end{tabular}
\caption{Dataset bias: training 2nd (and 3rd) layer(s) with data from another dataset. SUFR-W works better as the third layer because it has more people (400), while LFW only has about 150 people with more than 10 images. Model description: HOG features, \#eigen vector 1200, \#consensus 1500, Hash code length 28, Hash table number 20. }
\label{fig:dataset_bias}
\end{center}
\end{table}

\begin{table*}[t]
  \begin{center}
  \begin{tabular}{|cc||cc|p{3.3cm}c|}
 \hline  
\multicolumn{6}{|c|}{LFW \textbf{no outside data used} } \\
 \hline  
   \multicolumn{2}{|c||}{Aligned} &  \multicolumn{4}{c|}{Unaligned} \\
 \hline  
{\bf \scriptsize Model}  & {\bf Acc.}  & {\bf \scriptsize Model}  & {\bf Acc.} & {\bf \tiny Model (translation-invar.) }  & {\bf Acc.}    \\
 {\scriptsize Wolf et al. \cite{wolf2008descriptor} }& {\scriptsize 78.47\%}     & {\scriptsize Nowak et al. \cite{nowak2007learning} }& {\scriptsize 72.45}\% & {\tiny SIFT-BoW+SVM(Baseline) }&  {\scriptsize 57.73$\pm$2.53}\% \\
 {\scriptsize  V1-like/MKL \cite{pinto2009far} }& {\scriptsize 79.35}\%     & {\scriptsize  Sanderson et al. \cite{sanderson2009multi} } & {\scriptsize 72.95}\%     & {\tiny Our Model(HOG) }& {\scriptsize 84.73$\pm$1.82}\%\\
 {\scriptsize  APEM (fusion) \cite{cuifusing}} & {\scriptsize 84.08}\%    &  {\scriptsize  MRF-MLBP \cite{Arashloo2013MRFs} }&  {\scriptsize 79.08}\% & {\tiny Our Model(HOG+LBP)} & {\scriptsize  \textbf{86.15$\pm$1.50}\%}\\ 
 {\scriptsize  Simonyan et al. \cite{Simonyan2013} } & {\scriptsize 87.47}\%   & {\scriptsize  APEM (fusion) \cite{cuifusing} }& {\scriptsize 81.70}\% & {\tiny Our Model(HOG+LBP)+SVM} & {\scriptsize  \textbf{87.55$\pm$1.41}\%} \\
 \hline  
\end{tabular}
\caption{Our model (87.55$\pm$1.41\%) significantly outperforms state-of-the-art: APEM (81.70$\pm$1.78\%) in the LFW ``unaligned \& no outside data used'' category. The last column shows models that are translation-invariant. For the last model, we simply replace the final cosine distance classifier with a RBF-SVM, and we used the difference of the testing pair's third layer signatures as the input to the SVM.  }
\label{table:final_performance}
\end{center}
\end{table*}


\begin{table}[h!]
 \begin{center}
  \begin{tabular}{|c|cc|}
 \hline  
{\bf Model}  & {\bf LFW}  & {\bf LFW-J} \\
 \hline  
 HOG+SVM (baseline)  & 74.45/67.32\%     & 55.28\% \\
 Our Model (HOG)     & 84.73\%  & 84.62\% \\
 Our Model (HOG+LBP) & 86.15\%  & 86.02\% \\ 
 Our Model (HOG+LBP) + SVM & 87.55\%  & 87.45\% \\
  \hline  
\end{tabular}
\caption{The performance on LFW-original (unaligned) and LFW-J (jittered) datasets. For LFW-J: translation range: -40 to 40 pixels, scaling range: 1 to 1.5, in-plane rotation range -20 to 20 degree.  We used \textit{exactly} the same model for LFW and LFW-J. The  HOG baseline, $74.45\%$ uses the closely cropped performance and $67.32\%$ is non-cropped performance. With jittering, one cannot crop the image. Either way, HOG performance drops dramatically.}
\label{table:jittered_performance}
\end{center}
\end{table}

\vskip -.2in

\section{Conclusion}
What then is the answer to this paper's title question? Can the proposed network effectively replace DAR pipelines for unconstrained face recognition? The results are promising in that direction. Our system achieves one of the strongest reported accuracies in LFW's ``unaligned \& no outside data used'' category (fig. \ref{table:final_performance}). It also performs well on two more difficult datasets: SUFR-W and a significantly jittered (misaligned) version of LFW (example images in fig. \ref{fig:gallery}). 

Another encouraging result is the proposed system's robustness to the choice of dataset from which to obtain training templates (fig. \ref{fig:dataset_bias}). This suggests that its strong performance is unlikely to be attributable to overfitting. The finding that it is better to train on SUFR-W and test on LFW than vice-versa is likely due to the presence of non-frontal faces in the former dataset but not the latter. 

The proposed hierarchical approach has applications in computational neuroscience: 1. as a proof of principle that a feedforward network is able to do unconstrained face recognition, 2., as a starting point for developing models of the ventral stream and its face-specific branch\footnote{The feedforward network of face-specific patches  of visual cortex described by \cite{Tsao2006,freiwald2010functional}.} and 3., as a justification for the use of large templates to mitigate the problem of clutter---which is particularly interesting since large templates are known to provide an explanation for holistic face effects \cite{Tan2013} like the composite face effect \cite{young1987configurational}.   

We developed our system in order to address the question of whether a unified hierarchy could perform competitively with the dominant DAR approach. However, it is also interesting to consider the possibility of synergies between the two. Nothing about our method precludes its inclusion \textit{as} the recognition (or recognition + alignment) module within a standard DAR pipeline.  Such a hybrid system may be able to recover from errors in previous pipeline stages. For example, an invariant recognition module may be able to rescue a positive result despite an error in the alignment stage.  We think this is a promising avenue for future investigation.

\section*{Acknowledgments}
We thank Georgios Evangelopoulos for his helpful comments on an early version of this work. This material is based upon work supported by the Center for Minds, Brains and Machines (CBMM), funded by NSF STC award CCF-1231216.

\bibliographystyle{splncs}
\bibliography{My_mendeley_citations_in_bibtex}

\end{document}